\documentclass{article}

\usepackage{arxiv}

\usepackage[utf8]{inputenc} % allow utf-8 input
\usepackage[T1]{fontenc}    % use 8-bit T1 fonts
\usepackage{hyperref}       % hyperlinks
\usepackage{url}            % simple URL typesetting
\usepackage{booktabs}       % professional-quality tables
\usepackage{amsfonts}       % blackboard math symbols
\usepackage{nicefrac}       % compact symbols for 1/2, etc.
\usepackage{microtype}      % microtypography

\usepackage[english]{babel}
\usepackage{graphicx}
\usepackage[authoryear, round]{natbib}
\usepackage{lineno}
\usepackage{amsmath}

\DeclareMathOperator*{\argmin}{arg\,min}

\title{Predicting into unknown space? \\
Estimating the area of applicability of spatial prediction models}

\author{
  Hanna Meyer \\
  Institute of Landscape Ecology \\
  Westf{\"a}lische Wilhelms-Universit{\"a}t M{\"u}nster \\
  Heisenbergstr.2, 48149 M{\"u}nster, Germany \\
  \texttt{hanna.meyer@uni-muenster.de} \\
   \And
  Edzer Pebesma \\
  Institute for Geoinformatics \\
  Westf{\"a}lische Wilhelms-Universit{\"a}t M{\"u}nster \\
  Heisenbergstr.2, 48149 M{\"u}nster, Germany \\
  \texttt{edzer.pebesma@uni-muenster.de}
}

\begin{document}
\maketitle

\begin{abstract}

Predictive modelling using machine learning has become very popular for spatial mapping of the environment, even ambitiously on a global scale. Models are often applied to make predictions far beyond sampling locations where new geographic locations might considerably differ from the training data in their environmental properties. However, areas in the predictor space without support of training data are problematic. Since the model has no knowledge about these environments, predictions have to be considered highly uncertain. 

Estimating the area to which a prediction model can be reliably applied is required. Here, we suggest a methodology that delineates the "area of applicability" (AOA) that we define as the area, for which, on average, the cross-validation error of the model applies. We first propose a "dissimilarity index" (DI) that is based on the minimum distance to the training data in the multidimensional predictor space, with predictors being weighted by their respective importance in the model. The AOA is then derived by applying a threshold based on the DI of the training data where the DI of the training data is calculated with respect to the cross-validation strategy used for model training. We test for the ideal threshold by using a large set of simulated data and compare the prediction error within the AOA with the cross-validation error of the model.
Finally, we illustrate the approach in a simulated case study chosen to mimic ecological realities.

Our simulation study suggests a threshold on DI to define the AOA at the .95 quantile of the DI in the training data. Using this threshold, the prediction error (RMSE) within the AOA is comparable to the cross-validation RMSE of the trained model, while the cross-validation error does not apply outside the AOA. This applies to models being trained with randomly distributed training data, as well as when training data are clustered in space and where spatial cross-validation is applied.

Restricting predictions to the AOA is important especially when predictions are used for decision making or subsequent environmental modelling. We suggest to report the AOA alongside predictions, complementary to validation measures.

\end{abstract}

% keywords can be removed

\section{Introduction}
Spatial mapping is an important task in environmental science to reveal spatial (and spatio-temporal) patterns and changes of the environment. Predictive modelling is a common method in this context, where field data are used to train statistical models using spatially continuous predictor variables, e.g. derived from remote sensing imagery. The resulting model is then used to make predictions for the entire area of interest, i.e. beyond the locations of training data. In the last years, machine learning algorithms have become the popular tool in predictive modelling being able to capture nonlinear and complex relationships. In this way, a large variety of different environmental variables have been mapped even ambitiously on a global scale, such as global tree restoration potential \citep{Bastin2019}, soil properties \citep{Hengl2017}, distribution of nematodes \citep{Hoogen2019} and soil bacteria \citep{Delgado-Baquerizo2018}, global leaf-freezing resistance \citep{Zohner2020} or plant species Red List status \citep{Pelletier2018} to mention just a few.
However, after the reliability of global prediction maps has been frequently questioned, confidence in maps created this way is increasingly lost \citep[e.g. see comments to the highly discussed paper of][]{Bastin2019}.
Improved analysis and communication of uncertainties of spatial predictions is therefore required. This is important to identify locations where predictions are too uncertain to be considered for further action, e.g. in the context of prioritizing conservation assessment \citep{Pelletier2018} or if predictions are used as input for subsequent modelling  where a large error propagation should be avoided. 

Performances of machine learning models are typically communicated via (cross-) validation estimates. In the context of spatial mapping, the relevance of accounting for spatial dependencies and clustered sampling for reliable performance estimation has been recently highlighted by many studies \citep{Brenning2012a, Roberts2017, Schratz2019, Meyer2018, Valavi2018, Pohjankukka2017}. Spatial (cross-) validation allows giving a general error estimate for the predictions that is less sensitive to spatial dependence and spatially clustered training data than cross-validation based on random partitioning, however, we argue that this is not sufficient to communicate performance of prediction maps: typically field samples used as training data for predictive mapping are not evenly distributed over study areas and often predictions are made for areas that are lacking a support of training data. E.g. in the global map of soil nematode densities of \citet{Hoogen2019} central Africa as well as North East Asia is lacking any training data, but predictions are made for these areas. By transferring the model beyond the training locations (i.e. to new geographic space) it is assumed that the learned relationships between predictors and response still hold true. 
However, especially in heterogeneous landscapes, the new geographic space might differ considerably in its environmental properties from what has been observed in the training data. This leads to a question that is not addressed by (cross-) validation so far: what happens if the algorithm has never seen such environmental properties? This is relevant in so far as most machine learning algorithms can fit very complex relationships, but at the same time they are weak at extrapolation \citep{Qiao2019}. Therefore, gaps in the predictor space where there is no support of training data must be considered problematic because the algorithm has no knowledge about these environments and predictions for such areas must be regarded highly uncertain. 

Hence, there is a need to estimate the spatial areas for which the model is expected to give reliable predictions, which we will call the "area of applicability" (AOA) of a prediction model. Similar concepts have been discussed mainly in the field of chemical modelling \citep[Quantitative Structure-Activity Relationship models (QSAR), see e.g.][where the concept is usually referred to as "domain of applicability"]{Mathea2016, Toplak2014, Gadaleta2016} but they are only marginally addressed in the context of spatial predictive modelling \citep[e.g. in the context of soil mapping][]{Zhu2015}. Instead, models are often assumed to be applicable to the entire area of interest (e.g. globally). An exception in many global prediction maps is often Antarctica that is masked from the prediction maps \citep[e.g.][]{Hoogen2019} because, by expert knowledge, it is clearly outside of the AOA for most models. However, other environments might have to be considered equally unsuitable for model application. This can be very obvious (e.g. high mountain ranges using a model trained in lowlands) but also hard to assess by expert knowledge when areas feature combinations of environmental variables that are not covered by training data.

\begin{figure}[htbp]
\begin{center}
\includegraphics[width=0.8\textwidth]{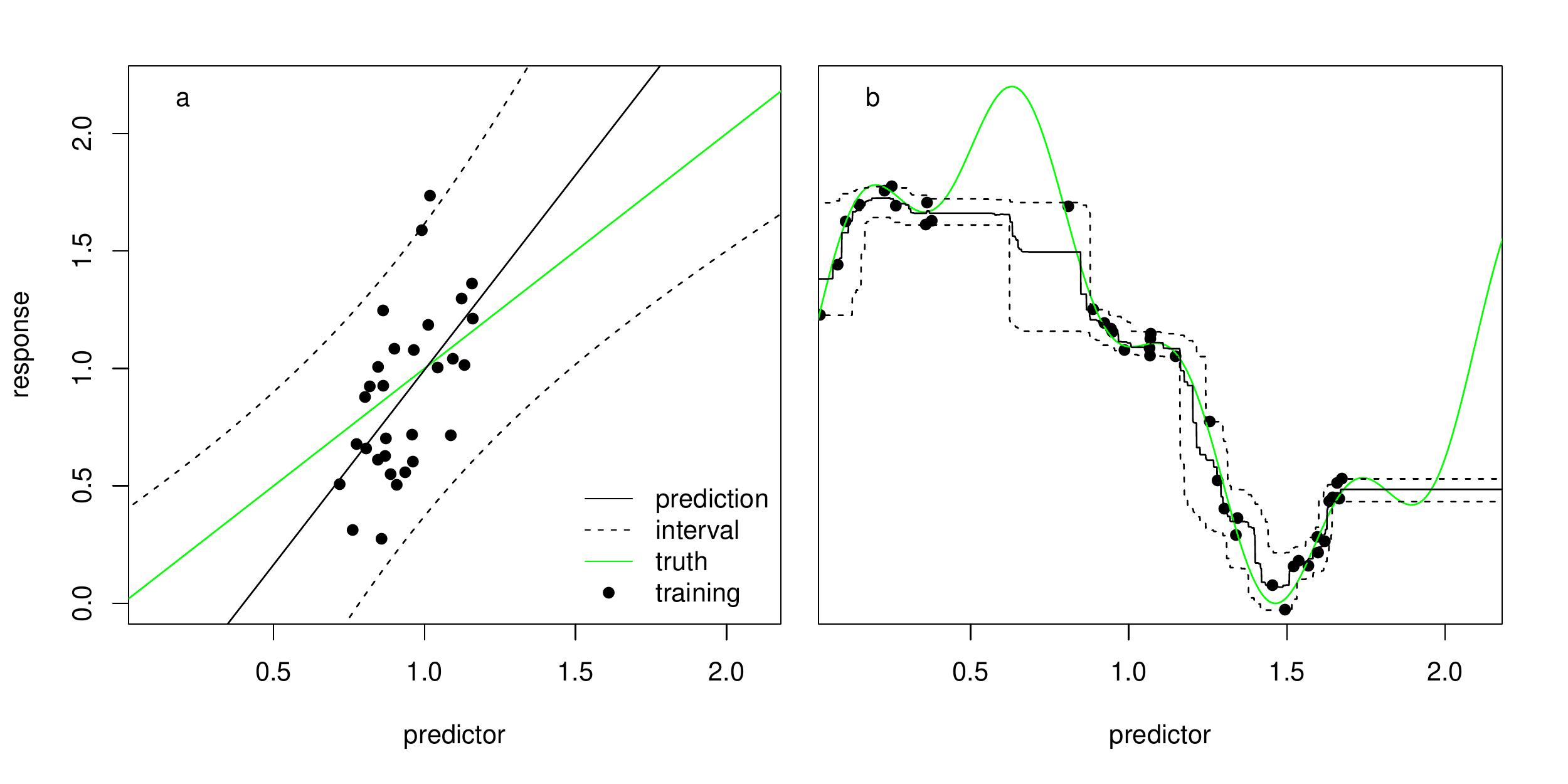}
\end{center}
\caption{Problem of predicting beyond the training data and behaviour of prediction intervals for different models. Left: linear regression prediction interval width increases with distance from the center of the training data,
right: a more complex relationship fitted with Random Forest. Prediction beyond the training data becomes highly unreliable, although prediction interval width outside the data range is constant.
Random Forest prediction intervals were obtained by computing quantiles over the predictions from individual trees.}
\label{fig:extrapolation}
\end{figure}

This aspect is not addressed by common approaches of uncertainty estimation in machine learning, which are usually based on variance of predictions made by ensembles of models. Hence, uncertainty estimates are derived from the variance of predictions made by individual models within the ensemble \citep[e.g. ][in the field of spatial mapping]{Coulston2016, Hoogen2019, Bastin2019}. This approach is very obvious for ensemble-based algorithms like Random Forests, where each tree is regarded as a model of an ensemble, and the variation in predictions among individual trees is used to quantify uncertainty (e.g. standard deviations of individual predictions or prediction intervals). Missing knowledge about environments is also not addressed by the quantile regression forests \citep{Meinshausen2006} that are occasionally suggested in the context of spatial mapping \citep{Hengl2018a,Vaysse2017} to derive prediction intervals based on the conditional distribution of the response variable.
Figure \ref{fig:extrapolation} shows an uncertainty estimation for a linear regression model and a Random Forest model. Clearly, both provide intervals that only make sense in the context of the respective models being valid, but it shows that Random Forest prediction intervals estimated from variability of predictions of the ensemble do not acknowledge that prediction gets harder when one moves further away from the training data, outside the data range or into significant gaps.
While these uncertainty estimates give valuable information on the variance in predictions and hence on locations where predictions are robust within the model, these approaches give no information about the AOA because dissimilarities in the predictor space between training and new data are not considered. 

To estimate the AOA, an obvious approach to account for dissimilarities is to look at distances in the predictor space between training data and a new data point \citep[e.g.][]{Sheridan2004}. Using raw predictor space distances may be problematic because in a machine learning model, typically certain variables have a high importance while others may be completely irrelevant (i.e. they differ in the degree to which they drive the prediction patterns). 
To approach this, \citet{Janet2019} suggest to use the distance in the latent predictor space of a neural network. However, this approach is specific to neural networks and not generically applicable to e.g. Random Forests.

Here we suggest a new method to map the AOA of spatial prediction models by accounting for the missing knowledge of the model about environments. The proposed measure is, in it's basic concept, based on distances in the predictor variable space. To account for varying importance of predictors, we compute distances by weighting predictors by their importance derived from the trained machine learning model. A dissimilarity index (DI) normalizes the minimum distance to a training data point by the average dissimilarity observed in the training data. The DI provides the basis to estimate the AOA that allows identifying areas where predictions should be regarded as highly uncertain. We define the AOA as the area for which, on average, the cross-validation error of the model applies. We estimate this area by using a quantile of the DI values observed in the training data as a threshold, where the DI values of the training data are calculated with respect to the cross-validation strategy being used for model training. Both the distances in the predictor space as well as the appropriate cross-validation strategy are affected by dependencies in the data intentionally or unintentionally associated with the sampling design. Therefore, we develop the method with the intention that it is applicable across sampling designs, and discuss the effect of sampling design and cross-validation strategy on AOA.  

%%%%%%%%%%%%%%%%%%%%%%%%%%%%%%%%%%%%%%%%%%%%%%%%%%%%%%%%%%%%%%%%%%%%
%%%%%%%%%%%%%%%%%%%%%%%%%%%%%%%%%%%%%%%%%%%%%%%%%%%%%%%%%%%%%%%%%%%%
\section{Methods}
We suggest a method that provides a unitless measure for expressesing how different a new data point is from the training data. We call this the "dissimilarity index" (DI), and suggest a threshold for it to define the "area of applicability" (AOA) of a model.
The method requires two datasets that are part of any supervised predictive modelling task. The first dataset is the training dataset that includes the sampling locations that are intended as training data used for model training. The second dataset contains the new locations where predictions should be made for. In the case of spatial mapping, this is the set of spatially continuous data, usually raster data with predictor variable values that are known for the entire area of interest. 

\subsection{Standardization of predictor variables}
To ensure that all variables are treated equally, the predictor variables are scaled by dividing mean-centered values by their respective standard deviations,
%\begin{equation}
%\label{eq:norm}
$$X^s_{i,j} = (X_{i,j} - \bar{X}_{\cdot,j})/\sigma_j$$
%\end{equation}
where $X^s_{i,j}$ refers to the scaled value of the $j$-th predictor variable corresponding to the $i$-th observation, 
$\bar{X}_{\cdot,j}$ to the mean and $\sigma_j$ to the standard deviation of the $j$-th predictor variable, and where 
mean and standard deviation are computed over the training data.

\subsection{Weighting of variables}

If distances were calculated based on the standardized predictors, all variables would be treated equally important. 
However, distances are not equally relevant within the predictor space but some variables are more important than others in the machine learning model and hence are mainly responsible for prediction patterns. Most machine learning models provide an estimate of relative variable importance \citep[see e.g. overview in][]{Kuhn2008}. 
To reflect the variable importance in the computation of distances in the predictor variable space,
we multiply the scaled variables with the importance estimate $w_j$ for each variable $j$ before distance calculation takes place, 

%\begin{equation}
$$X^{sw}_{j} = w_j X^{s}_{j}.$$
%\label{eq:weights}
%\end{equation}

As a consequence, distances in the predictor space in the direction of the more important variables have a higher effect on our dissimilarity measure.

\subsection{Multivariate distance calculation}
The Euclidean distance between two arbitrary points $a$ and $b$ in the predictor variable space is calculated as 
%\begin{equation}
$$d(a,b) = \sqrt{ \sum_{j=1}^{p} (X^{sw}_{a,j} - X^{sw}_{b,j})^2}.$$
%\label{eq:dist}
%\end{equation}

\begin{figure}[htbp]
\includegraphics[width=0.6\textwidth]{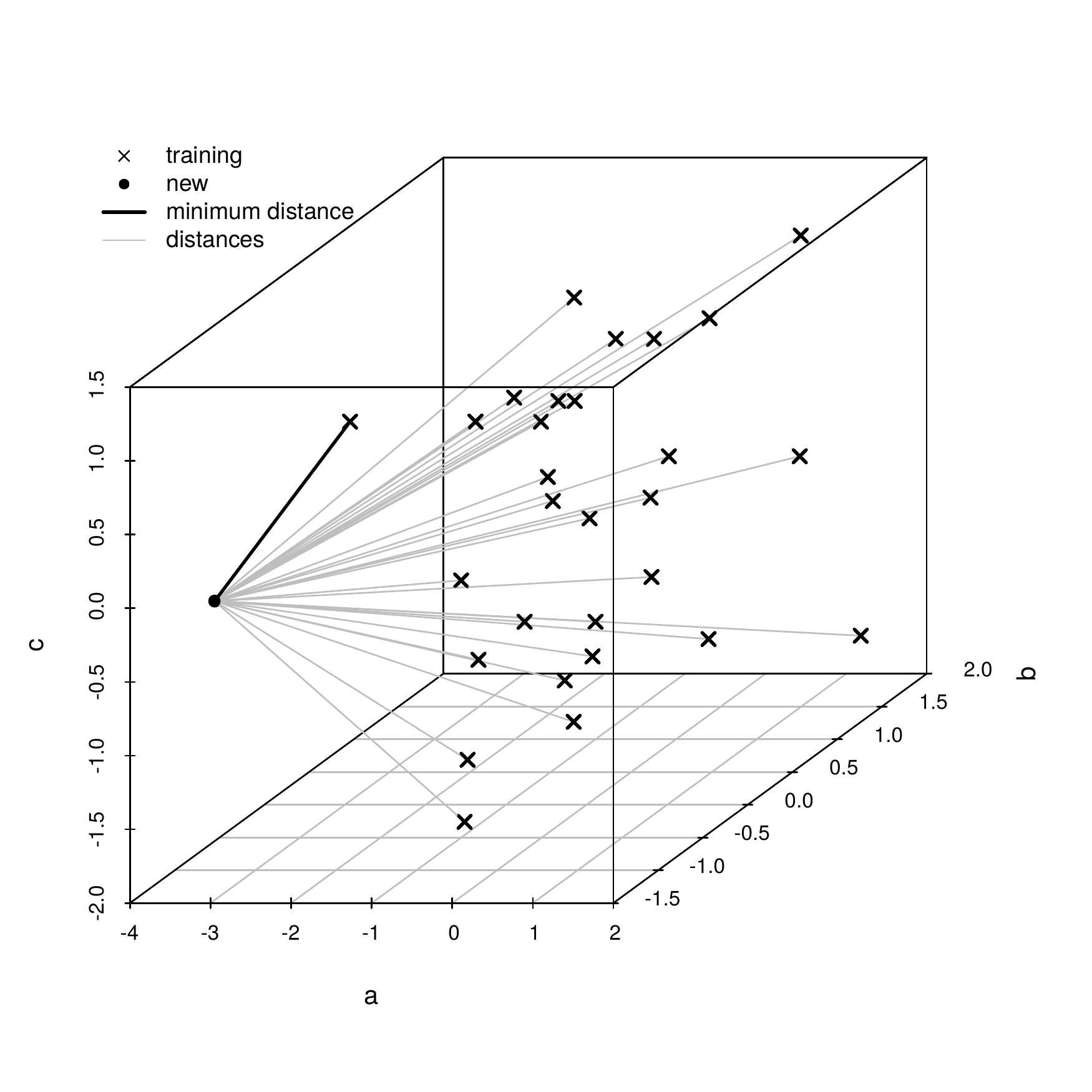}
\caption{Example of how the dissimilarity index (DI) for a new data point is calculated: The distance in the normalized and weighted predictor space (here indicated by three virtual predictor variables a, b, c) to each training data point is calculated (grey lines) to identify the closest data point. The distance to the closest data point (black line) is then divided by the average of the mean distances between the training data.
Here the average mean distance between training data is 2.29, the distance to the closest training point is 2.46. Hence the DI is 1.07.}
\label{fig:3dscatter} 
\end{figure}

For a new prediction location $k$, the distance to the nearest training data point $i$

$$d_k = \argmin_i d(k,i)$$

is used to calculate the DI (Figure \ref{fig:3dscatter}). 

\subsection{Dissimilarity index}
\label{sec:DI}
To allow for interpretation and comparison between models, we standardise distances in predictor space for new prediction locations $k$ by dividing the minimum distance $d_k$ by the average of the distances in the training data $\bar{d}$, and call this the dissimilarity index $\mbox{DI}_k$, defined as

%\begin{equation}
$$\mbox{DI}_k = d_k/\bar{d}$$
%\label{eq:index}
%\end{equation}
with $\bar{d}$ the average of all pairwise distances between the $n$ training data.

Using the standardised weighted distances, DI can take values ranging from 0 to $\infty$.
If the result is 0, the new data point is identical in its predictor properties to a training data point. With increasing values of the DI the distance to a nearest training data point increases. If the values are greater 1, the difference to the nearest training data point is larger than the average dissimilarity (i.e. average distance) between all training data pairs. 

\subsection{Deriving the area of applicability}
\label{sec:threshold}
The AOA is the area for which we expect the model can make reliable predictions. Reliable predictions are defined here as predictions that can be made with an error that is, on average, comparable to the cross-validation error of the model computed over the training data.
We choose the AOA as the area where the DI does not exceed a certain threshold. With regard to our definition that the AOA is the area for which, on average, the cross-validation error applies, the threshold is derived from the DI values of the training data, with distances calculated between points that do not occur in the same cross-validation fold.
Taking the cross-validation folds into account is required here because cross-validation is based on repeatedly leaving training data out, hence we assume the model error applies to areas that have a comparable DI as the DI of the cross-validated training data. Therefore, we calculate the DI for each training data point as described in section \ref{sec:DI}, however, in line with the cross-validation strategy being used, distance is measured to the nearest training data point (in the predictor space) that is not in the same cross-validation fold. E.g. if a 10-fold random cross-validation is applied for model validation, data are randomly split into 10 folds and the cross-validation error is the average over the prediction errors for each of the folds held back for validation. Hence, the predictions for a respective fold are based on a model being trained on the remaining data. In line with that, the DI for each training data point is calculated by using the the distance to the nearest training data point that is not in the same fold.

To choose a threshold based on the DI of the cross-validated training data, we tested varying thresholds for nearly 1000 prediction tasks where the true values were known. As prediction tasks, we used a spatially continuous response variable of Europe that is simulated based on bioclimatic predictor variables (\url{www.worldclim.org/bioclim}) and the simulation approach of \cite{Leroy2016}. This approach was developed as an example of virtual species suitability but will be used here as an arbitrary response variable based on environmental predictors.
The application of an area-wide simulated response variable is important here, as it allows to compare the predictions with true values, and based on that comparison decide which threshold for DI defines an AOA.

As predictor variables, we used Worldclim \citep{Hijmans2005} to choose 19 bioclimatic variables. The response variable was generated by a principal component analysis (PCA) of a subset of the bioclimatic variables. The variables used to simulate the response were the mean diurnal range ("bio2"), maximum temperature of the warmest month ("bio5"), mean temperature of the warmest quarter ("bio10"), precipitation of the wettest month ("bio13"), precipitation of the driest month ("bio14"), and precipitation of the coldest quarter ("bio19"). For the PCA the response to each of the first two principal components (axes) is defined and combined to create the final response variable. Therefore, the response to the two first axes of the PCA is determined with Gaussian functions as described in \citet{Leroy2016}. The means of the Gaussian response functions to the axes of the PCA was varied here between 1 and 3 (first axis) and -1 and 1 (second axis) and the standard deviations were varied between 1 and 3 for both axes, resulting in 81 different response variables.

The simulated response variable is available in a spatially continuous way, however, to simulate a typical prediction task, we simulated field sampling designs. We selected sample point locations randomly from the target area with varying sample sizes (n = 25, 50, 75, 100). Each combination of response variables and sample size was tested with three independent replicates of the random sampling design, resulting in total in $81 \times 4 \times 3=972$ different simulated "realities". 

We used Random Forests \citep{Breiman2001} as machine learning algorithm because it is one of the most frequently used algorithms in the context of environmental mapping \citep[e.g. used in the context of global mapping in][]{Bastin2019, Hengl2017, Hoogen2019}. To prepare model training, predictors and response were extracted for the locations of the sampling data points. For model training, the Random Forest implementation of \citet{Liaw2002} was used and accessed via the caret package \citep{Kuhn2019} in R \citep{RCT2020}. Each forest consisted of 500 trees and the number of randomly selected variables at each split (mtry) was tuned between 2 and 19 (the number of predictor variables). Tuning and performance estimation was done using random 10-fold cross-validation. The trained models were applied to the complete predictor variables to make spatial predictions over the entire area.
To assess the relative variable importance required for the estimation of the DI, the approach of \citet{Liaw2002} was used to estimate $w_j$. Importance is indicated as the increase of the mean squared error when a variable is randomly permuted. Hence, the higher the decrease, the higher the importance in the model.

From the catalogue of simulations, we computed prediction errors by subtracting predicted from true values, for all prediction locations. We then evaluated a number of candidate cutoff values for DI that could be used to determine the AOA, and chose the .25, .50, .90, .95, .99 and 1. quantiles of the DI values for training data points calculated by taking the cross-validation folds into account. For each quantile, the corresponding cutoff value allows to compare the RMSPE (differences between predicted and true values) with the RMSE (cross validation error of the model), in order to arrive at a quantile value of DI values for which RMSPE and RMSE correspond, on average, over the set of simulations.

%%%%%%%%%%%%%%%%%%%%%%%%%%%%%%%%%%%%%%%%%%%%%%%%%%%%%%%%%%%%%%%%%%%%
%%%%%%%%%%%%%%%%%%%%%%%%%%%%%%%%%%%%%%%%%%%%%%%%%%%%%%%%%%%%%%%%%%%%
\section{Results}
\subsection{Threshold estimation for the AOA}

\begin{figure}[htbp]
\includegraphics[width=0.75\textwidth]{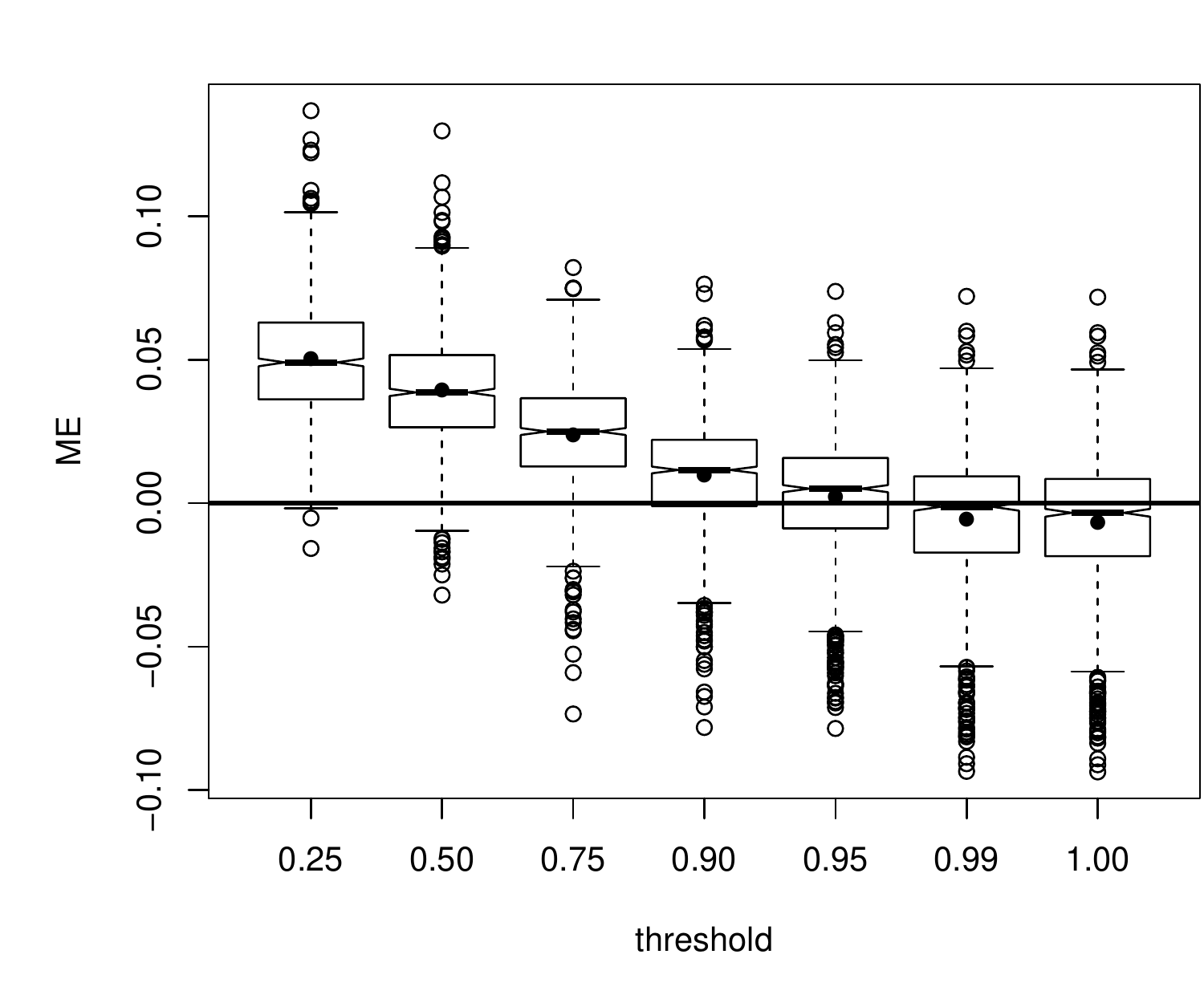}
\caption{Box-and-whisker plot of differences between the cross-validation RMSE and the RMSPE in the AOA over the 972 simulations, for a number of chosen DI thresholds for the AOA (quantiles of the DI values in the training data calculated using the minimum distance to other training data not occurring in the same cross-validation fold). Mean values are indicated with a solid bullet. At the .95 quantile threshold, the average of the RMSPE equals the RMSE (mean difference nearly zero), which suggests to choose this as the threshold to define the AOA.}
\label{fig:beyondCaseStudy_threshold} 
\end{figure}

\begin{figure}[htbp]
\includegraphics[width=1\textwidth]{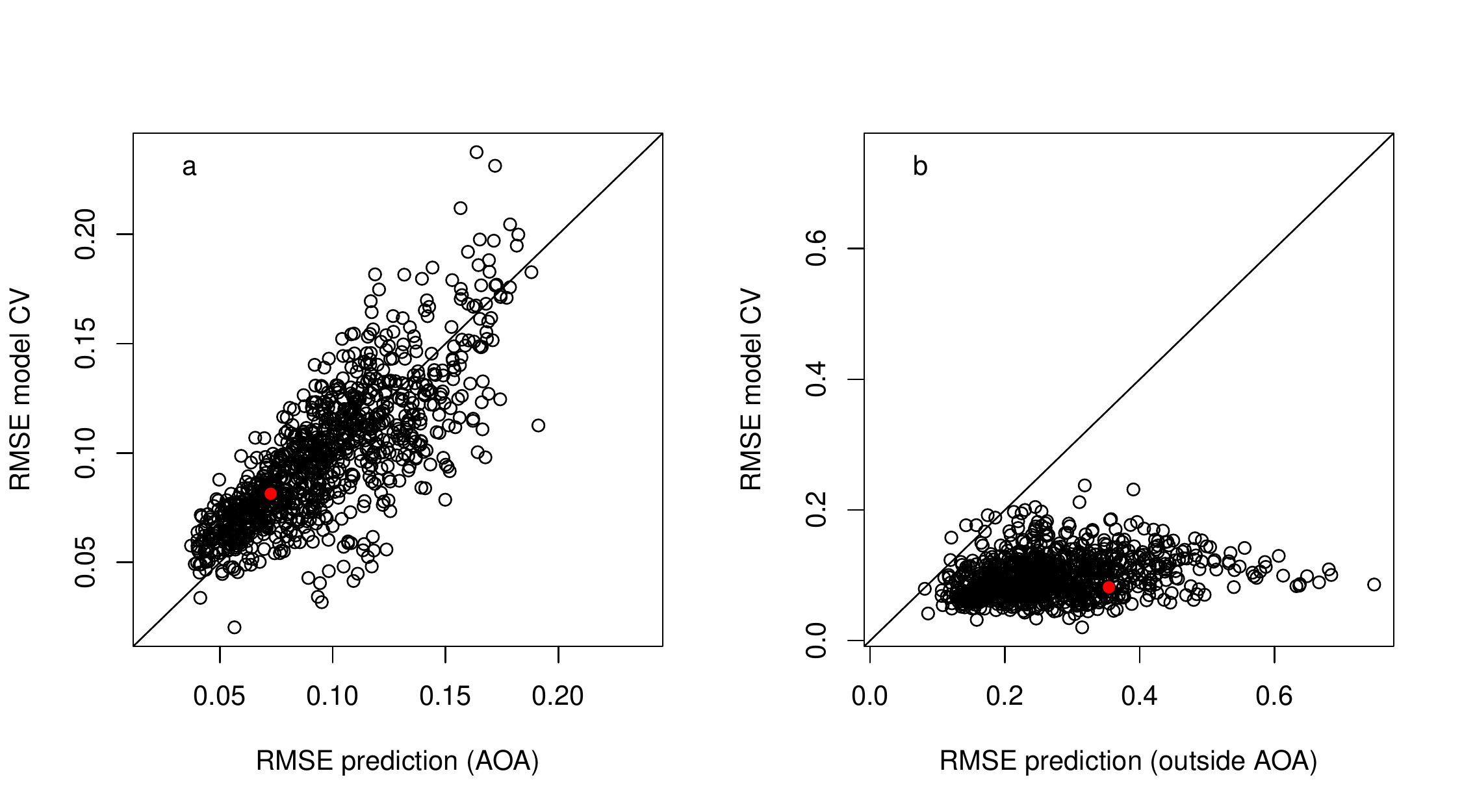}
\caption{RMSE of the model (cross-validation, y-axis) against RMSPE (true predicition errors outside training data, x-axis) for the 972 simulations, inside the Area Of Applicability (AOA; left) and outside the AOA (right); inside the AOA both errors correspond on average, outside this area the RMSPE is much larger.}
\label{fig:beyondCaseStudy} 
\end{figure}

Based on the 972 model scenarios, Figure \ref{fig:beyondCaseStudy_threshold} shows that the selection of the .95 quantile of the DI observed in the cross-validated training data represents a suitable threshold on the DI: the average mean difference between the cross-validation error and the prediction error was close to zero (0.002).
When the .95 quantile is applied as a threshold to the DI of the 972 case study scenarios, Figure \ref{fig:beyondCaseStudy}a shows that the prediction error within the AOA is in high agreement with the cross-validation error of the model. The model error is not valid outside of the AOA, indicated by considerably higher RMSE values for the prediction compared to the cross-validation error (Figure \ref{fig:beyondCaseStudy}b).

\subsection{Case Study}
\begin{figure}[htbp]
\includegraphics[width=1\textwidth]{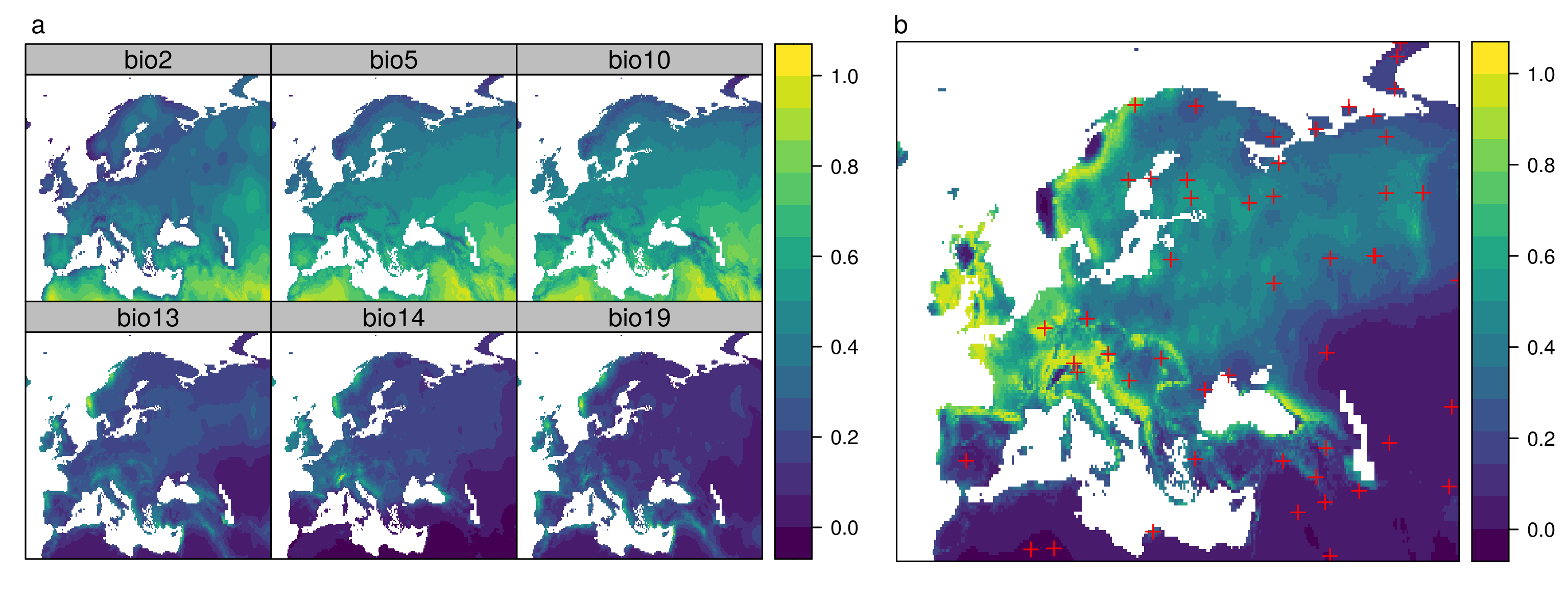}
\caption{Bioclimatic variables used to simulate the response variable for the case study. The variables are stretched here between 0 and 1 for visualization purposes (a). b shows the simulated response variable for the case study and the location of the 50 randomly selected sampling points (red markers) used as training data.}
\label{fig:trainingdata} 
\end{figure}

\begin{figure}[htbp]
\includegraphics[width=1\textwidth]{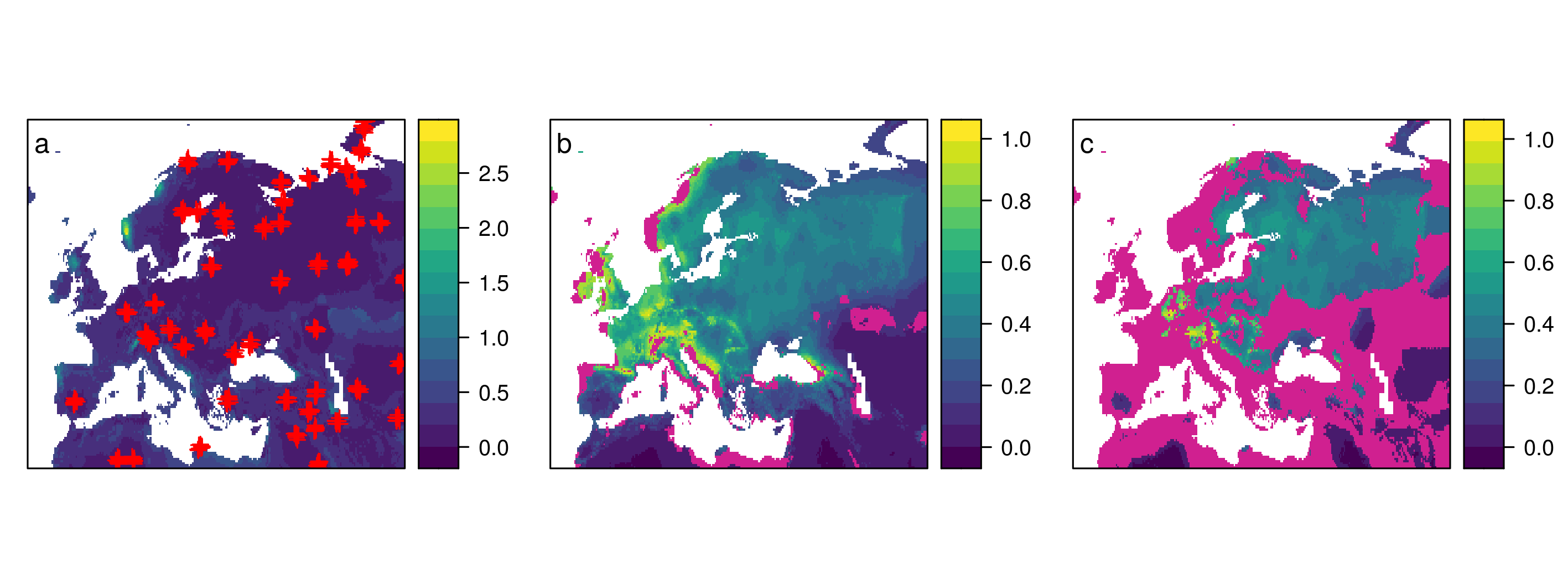}
\caption{Example of a clustered sampling design and the consequences for the estimation of the area of applicability: the dissimilarity index overlaid by 500 training data points clustered around 50 locations (a), predictions for the derived area of applicability for which the spatial cross-validation error applies (b), as well as predictions for the area of applicability for which the (in this case inappropriate) random cross-validation error applies. Areas outside the area of applicability are shown in pink.}
\label{fig:SpVsRandom} 
\end{figure}

To motivate our proposal, we show the application of the approach in an illustrated case study using a single simulation from the scenarios described above. Therefore we use a single setting from the 972 simulations where the response is developed from the bioclimatic predictors \ref{fig:trainingdata}a with means of the Gaussian response functions of 3 (first axis) and -1 (second axis) and standard deviations of 2 for both axes. The simulated response had values between 0 and 1 with a mean of 0.31 (Figure \ref{fig:trainingdata}b). As training data, we randomly selected 50 sample point locations from the target area (red markers in Figure \ref{fig:trainingdata}b).
To further illustrate the suitability of the presented methodology across sampling designs (and hence across suitable cross-validation strategies), as a second example, we simulated a spatially clustered sampling design: instead of 50 randomly selected sampling locations as described before, we simulated 10 sampling points clustered around each of the 50 locations, resulting in 500 training points across the 50 independent locations (Figure \ref{fig:SpVsRandom}a). Model performance estimation was done using random 10-fold cross-validation for the randomly distributed training data, as well as a leave-cluster-out spatial cross-validation for the clustered training data.

\begin{figure}[htbp]
\includegraphics[width=0.6\textwidth]{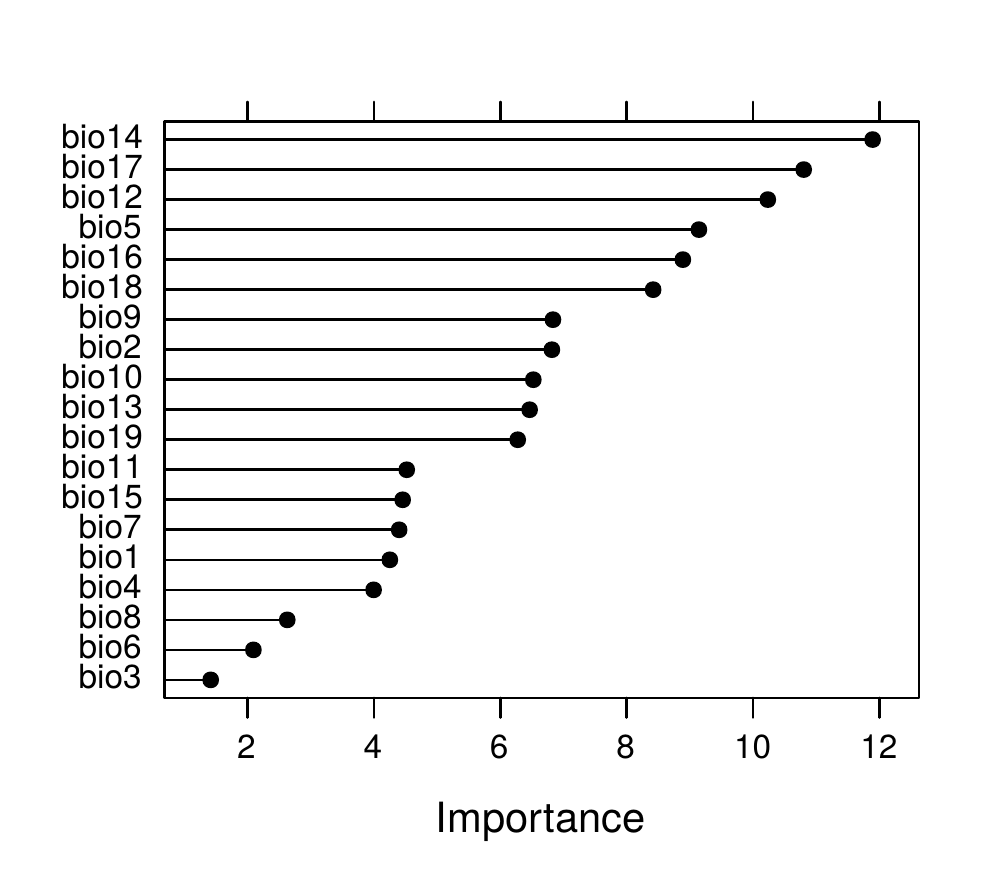}
\caption{Importance of the predictor variables within the Random Forest model based on the 50 randomly distribute sample data. Importance is indicated as the increase of the mean squared error when a variable is randomly permuted.}
\label{fig:varimp} 
\end{figure}

The importance of the different predictor variables ranged from 1.5 to 12 (Figure \ref{fig:varimp}) which represented the baseline for variable weighting used to estimate the DI. The AOA was estimated using the .95 quantile of the DI of the training data as a threshold, where the DI of the training data was calculated with consideration of the respective cross-validation strategy as described in section \ref{sec:threshold}. 
To highlight the advantage of the newly developed method, the DI was further compared to the commonly applied standard deviation of the Random Forest ensemble. Therefore, the standard deviations of the individual predictions made by the 500 trees were calculated for a respective pixel.

\begin{figure}[htbp]
\includegraphics[width=1\textwidth]{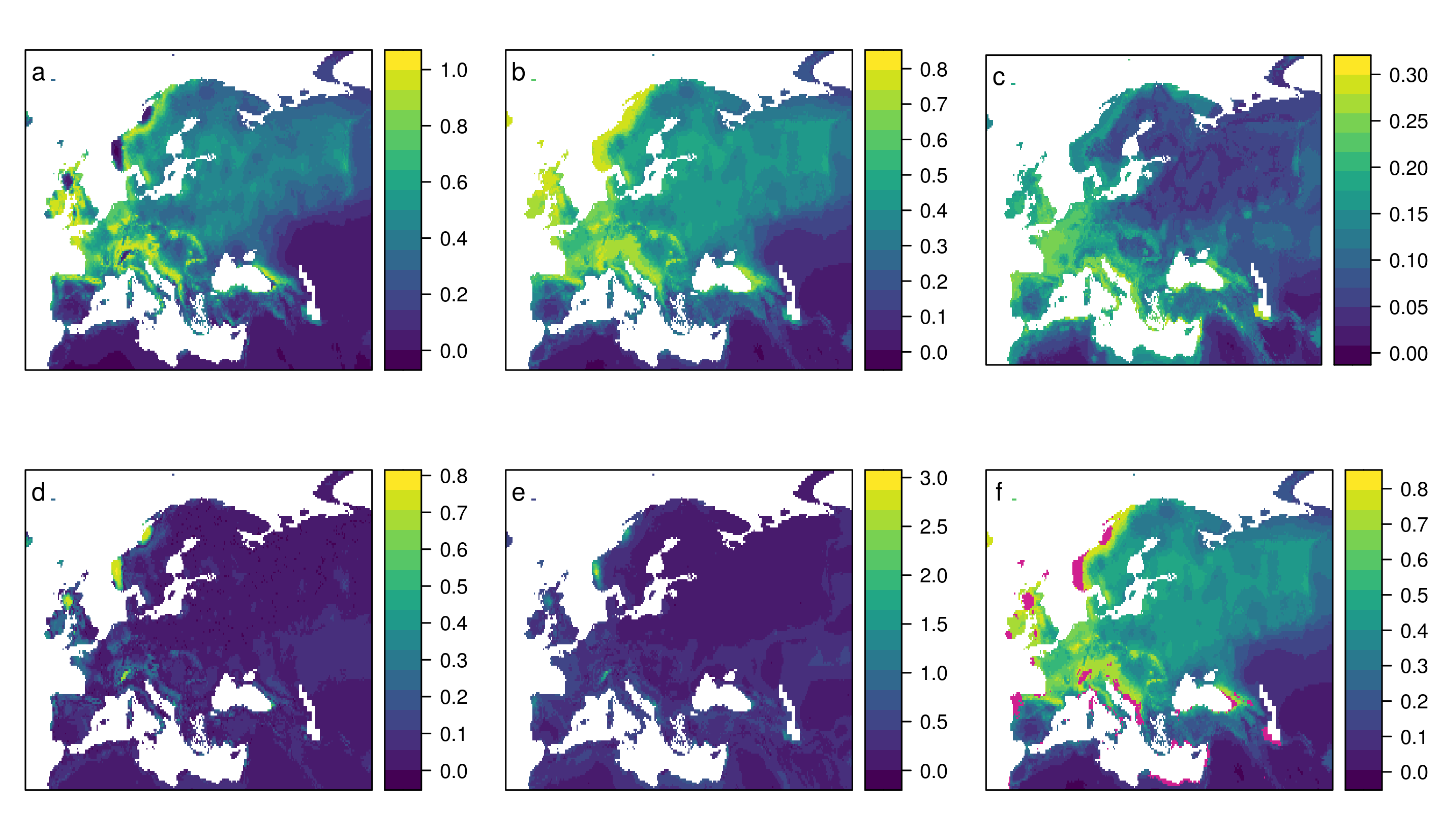}
\caption{Comparison between reference (a), prediction (b), standard deviation of predictions (c), the true absolute prediction error (d), the newly suggested dissimilarity index (e), and the predictions masked by the derived area of applicability where areas outside of the area of applicability are shown in pink (f).}
\label{fig:comp} 
\end{figure}

The results show that the case study model had a high ability to predict the response variable, indicated by a random cross-validation $R^2$ of 0.95 and a RMSE of 0.08 for the prediction task using randomly distributed data.
The DI (Figure \ref{fig:comp}e) shows clear spatial patterns across Europe. Values range from 0 to 2.89 with an average of 0.25. Noticeable are low values (low applicability) in the Alps and at the west coast of Norway. This means that these areas feature very distinct environments compared to the environments covered by the training data.

\begin{figure}[htbp]
\includegraphics[width=0.5\textwidth]{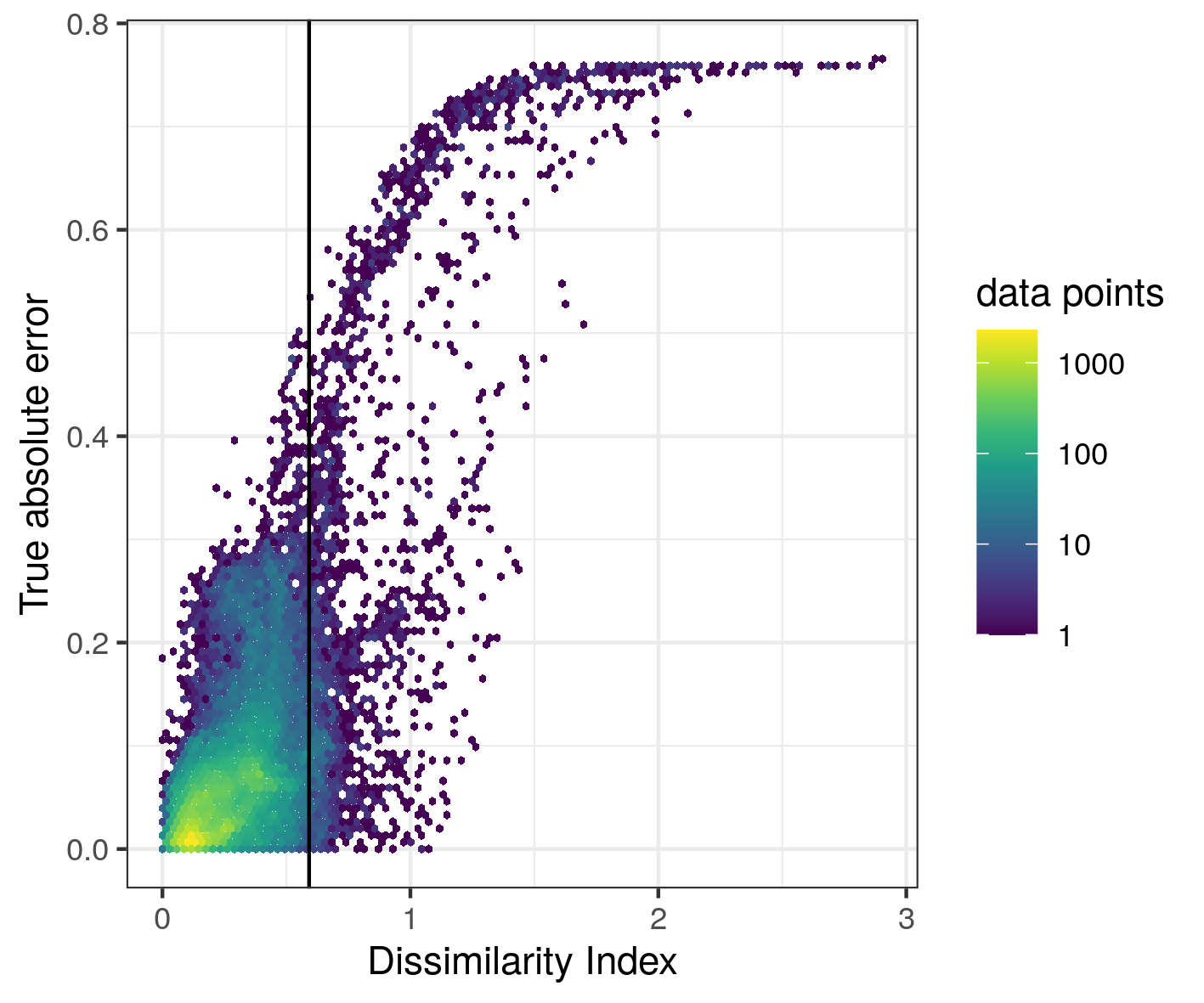}
\caption{Relationship between values of the dissimilarity index (DI, x-axis) and the true absolute error of predictions (y-axis). The colors represent the data point density. The vertical line represents the threshold used for the area of applicability.}
\label{fig:compHex} 
\end{figure}

The standard deviations of the Random Forest predictions feature very different spatial patterns (Figure \ref{fig:comp}c) that are not in agreement with the true absolute prediction error (Figure \ref{fig:comp}d). 
In contrast, the DI (Figure \ref{fig:comp}e) reflects the spatial patterns in the true error (Figure \ref{fig:comp}d), with a correlation coefficient of r=0.71 (Figure \ref{fig:compHex}). If variables were not weighted according to their relevance in the model (Figure \ref{fig:varimp}) the DI would be less in accordance with the true absolute error (r = 0.62).

The threshold for the AOA as derived by the .95 quantile of the DI observed in the training data was 0.59. Figure \ref{fig:comp}f shows the predictions made by the model (Figure \ref{fig:comp}b) but masked by the AOA. The average agreement between the reference and the predictions is higher within the AOA (r=0.97, RMSE=0.07) compared to the entire study area (r=0.92, RMSE=0.10). Outside the AOA, the agreement is considerably lower (r=0.16, RMSE=0.38). Note that the prediction error within the AOA is in high agreement with the random cross-validation error of the model (RMSE=0.08).

Using the spatially clustered data points for model training, the random cross-validation RMSE is 0.021. When validated with a leave-cluster-out spatial cross-validation, the RMSE increases to 0.039. 
Using the threshold on the DI estimated by taking into account distances to data points not located in the same spatial cluster, the AOA for which the spatial cross-validation applies (Figure \ref{fig:SpVsRandom}b) is considerably larger compared to the AOA for which the random cross-validation error applies (Figure \ref{fig:SpVsRandom}c). The true prediction RMSE within the AOA is in both cases comparable to the respective cross-validation RMSE: 0.043 for the AOA of the spatial model, and 0.024 for the AOA of the random model.

%%%%%%%%%%%%%%%%%%%%%%%%%%%%%%%%%%%%%%%%%%%%%%%%%%%%%%%%%%%%%%%%%%%%
%%%%%%%%%%%%%%%%%%%%%%%%%%%%%%%%%%%%%%%%%%%%%%%%%%%%%%%%%%%%%%%%%%%%
\section{Discussion}
\label{sec:discussion}

We propose a method to estimate the Area of Applicability (AOA) of predictive models, with which we mean the area where a model can be expected to make predictions with an average error that is comparable to the model error estimated using cross-validation. 

The AOA is found by thresholding the Dissimilarity Index (DI), a standardized distance in the multidimensional predictor space, using a quantile of the DI values of the training data.
Based on a catalogue of 972 simulations, we found that a threshold value equal to the .95 quantile of the DI values of the training data, calculated by considering the cross-validation folds, yielded an AOA with prediction errors that are on average similar to the cross validation error. 
Using this threshold, a new data point is outside of the AOA where DI exceeds the .95 quantile of the DI values observed in the cross-validated training data. The cross-validation error of the model should not be considered valid outside of the AOA because the DI (i.e. dissimilarity) is greater that the DI of the data used for model error estimation via cross-validation.

Knowledge on the AOA is especially relevant when predictions are made for heterogeneous areas but based on limited field data, or are even made across study areas where it is unclear if the model can be applied to the new environment. \citet{Yates2018} raised the need for assessing the transferability of prediction models as an "outstanding challenge". The method to estimate the AOA as presented here provides one suggestion to assess the transferability by quantifying the differences in the environmental conditions between training data and target area to identify the area for which the model can be expected to make reliable predictions (i.e. predictions with an error comparable to the communicated model performance).
Limiting predictions to the AOA is especially important when predictions, along with cross-validation estimates, are used as a baseline for decision making (e.g. in the context of nature conservation). It is further of high relevance when a prediction map is used for subsequent modelling to limit propagation of massive errors. E.g. the frequently used global soil maps of \citet{Hengl2017} represent a basis for many subsequent environmental prediction models \citep[e.g. for mapping soil organisms in][]{Hoogen2019, Delgado-Baquerizo2018}. The presented approach would allow assessing the spatial suitability of these products, and allow for constraining further application to the estimated AOA.

The AOA adds an important information to the validation metrics based on test data taken from the biased sample, or cross-validation strategies. First, the model validation usually provides a global estimate that does not allow for representing the varying performance of the model in a spatial way.
Second, (cross-) validation estimates are based on the sampling data only, but spatial sampling is usually biased \citep[e.g.][]{Kadmon2004, Bystriakova2012} and is unlikely to cover the entire environmental conditions of heterogeneous environments even when a sampling design is planned with the aim to cover the range of the predictor space \citep{Hengl2003}. Subsequently the validation that is based on independent subsets of the samples will be biased as well. One might argue that this is rather a problem of sampling strategies that need to be improved in the first place rather than addressing this issue by mapping the AOA of the model. However, the idea of machine learning for spatial mapping is that we are able to deal with complex relationships and a large variety of potential predictor variables. This challenges sampling by expert knowledge because gaps in the predictor space are hard to identify in high-dimensional predictor spaces. Also, in many prediction tasks data from large composite databases are used that lack a common or shared sampling design, or for which the sampling design is unknown. 

The uncertainty originating from missing knowledge about environments is also not reflected by standard deviations of predictions made by individual predictions of an ensemble. They give valuable insights into the model by indicating areas where the model is very sensitive to changes in the data or randomness, but do not provide information on the AOA. The results showed that it is required to account for incomplete coverage of environmental properties and to limit predictions to areas that are similar in their predictor properties compared to the training data and are hence within the AOA.

The AOA, by it's definition, depends on the cross-validation strategy, and the choice of the appropriate cross-validation strategy depends on the distribution of the training data. Sampling strategies can follow different designs and training data might come, intended or accidentally, clustered in both geographical and predictor space \citep[e.g. training polygons for land cover classifications, see][]{Meyer2019b}. For spatially clustered data, the "default" random cross-validation error is not meaningful, because it indicates the ability of the model to reproduce the within-cluster variability in the training data rather than the ability to make predictions beyond the data used for model training, leading to overly optimistic performance estimates \citep[e.g.][]{Meyer2019b,Meyer2018,Roberts2017,Valavi2018,Pohjankukka2017,Schratz2019}.
Not only the cross-validation error, but also the strategy to estimate the AOA is highly sensitive to spatial dependencies in the data. The AOA is derived from the DI observed in the training data considering cross-validation folds. If (spatial) folds were not considered in the estimation of the DI in the training data, the minimum distance to any training data point would be used, which is not meaningful: values for the DI would be very low because of nearly identical neighbors, leading to an AOA restricted to the close proximity of the training data. This would be reasonable for a random cross-validation error, assuming that the overly optimistic cross-validation error should only be valid for that restricted area as well. However, as outlined above, the random cross-validation error is not meaningful but spatial cross-validation is required in this case. We therefore suggested that, in line with the appropriate cross-validation strategy being used \citep[ e.g. spatial, see][]{Meyer2018,Roberts2017,Valavi2018,Pohjankukka2017,Schratz2019}, the threshold for the AOA is derived from the training data, by calculating the minimum distance to other training data points that do not occur in the same cross validation fold. In this way the methodology to estimate the AOA is applicable across cross-validation strategies and should reflect the AOA for which the respective cross-validation error applies.

\begin{figure}[htbp]
\includegraphics[width=0.75\textwidth]{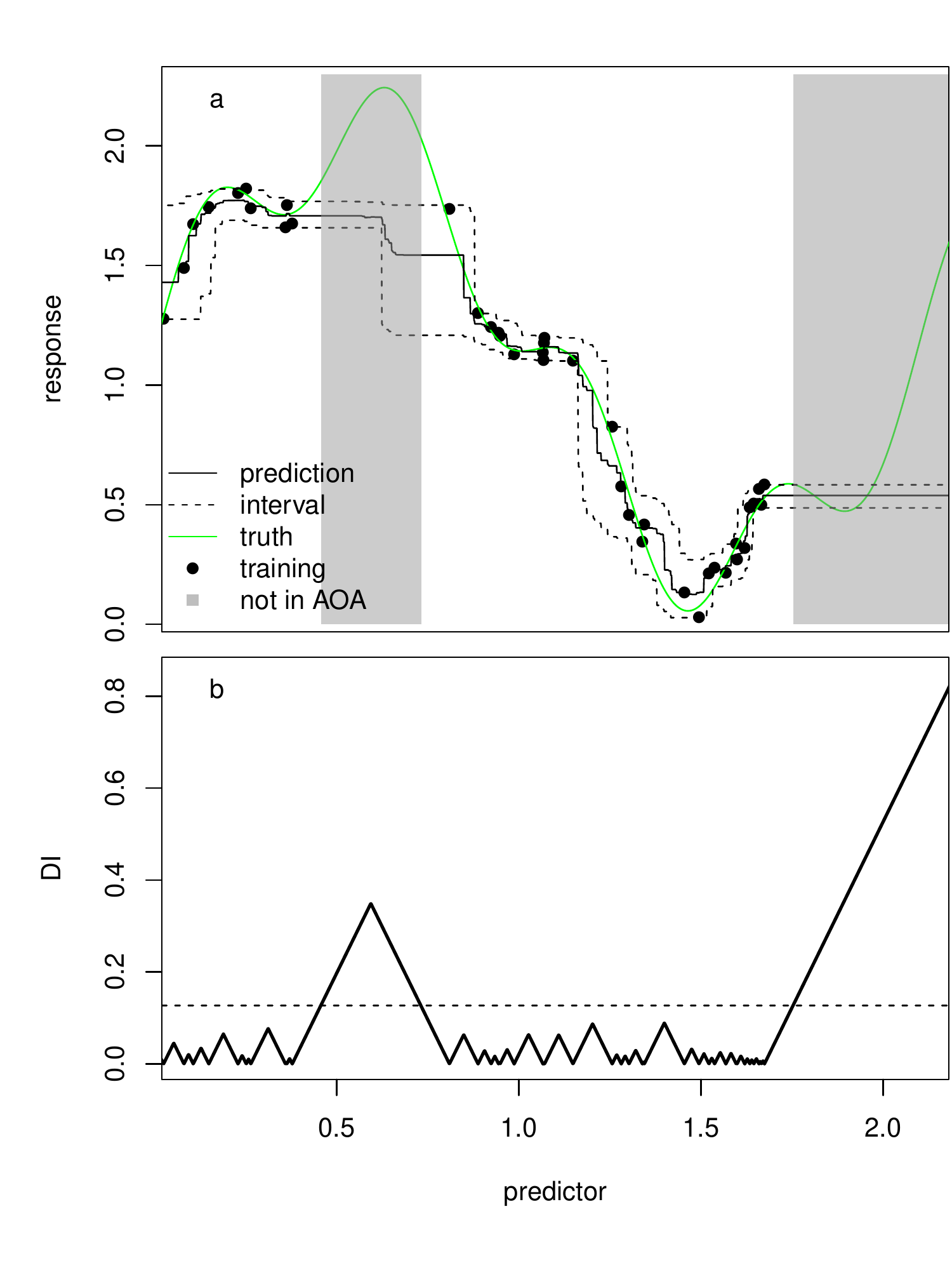}
\caption{Hypothetical example of a relationship between a virtual predictor and response variable as well as the predictions made by Random Forest already presented in Figure~\ref{fig:extrapolation} (a). Low values in the dissimilarity index (DI) do not necessarily mean that the prediction error is high. As moving away from the last training data point (x= 1.7), the value of the DI increases (b). However, the error does not necessarily increases in the same way (comparing the predictions with the truth in a). The uncertainty must still be considered as very high because this area of the predictor space is unknown to the model. The area of applicability (AOA) is derived using the .95 quantile of the DI observed in the training data as a threshold (dashed line in b) and is used to exclude predictions in areas where dissimilarity is too high (grey area in a).}
\label{fig:errorRel} 
\end{figure}

The application of simulated response variables was required here, to allow for an estimation of the true prediction error outside the training data. Note that the simulations applied here lead to very strong prediction models where the simulated response is a clear function of the predictors. Therefore, prediction errors can, to a large degree, be traced back to missing coverage in the environmental predictors. The relationship between the DI and a true error will be less strong for weak prediction tasks because missing knowledge of the environment will not be the major source of uncertainty. Other factors, like especially a poor ability of the predictors to model the response, influence uncertainty as well but are not considered in the DI, but by the (cross-) validation error and hence also by the AOA. Also note that high differences between training data and new data do not necessarily lead to a high prediction error (see Figure \ref{fig:errorRel}). Instead, locations with a high DI, falling outside the AOA, are associated with a high uncertainty because the environment, and hence the prediction success is unknown. 

The DI and the derived AOA do not only provide relevant information for assessing the reliability of predictions but can also serve model improvement.
The uncertainty originating from missing knowledge of the model represents a reducible part of the total prediction uncertainties, because it is based on the training data availability. Knowledge of the AOA allows to improve the model quality by targeting subsequent sampling effort to improve the data basis. Therefore, the suggested DI can be used to detect environments that are not covered by training data and hence can be used as a baseline for further sampling campaigns with the aim to minimize uncertainty in the predictions, and hence to increase the AOA of a model.
Since the estimation of the AOA requires the predictive model for variable weighting only, the effect of new samples on the AOA can be assessed without high computation times. It can also be an option that variable weighting is done by expert elicitation prior to a modelling procedure so that the approach can be deployed in the early stages of a research project starting with the selection of sampling locations.

The method to estimate the AOA as presented here should be considered a first attempt and contains a number of aspects that are up for discussion. These include:
\begin{itemize}
\item the use of distances in a weighted predictor space; weighting effectively alleviates the curse of dimensionality, but lacks a formal statistical argument,
\item the use of Euclidean distance; monotone transformation (e.g. log, or power transforms) of predictors would not affect a Random Forest model fit or prediction, but would strongly affect the AOA,
\item the use of the nearest training data point $d_k$; this does not discriminate between cases where one isolated, remote training point is nearest, or a predictor space location is surrounded by training points at this same distance; as an alternative, distances to multiple points (k-NN) could be used \citep[e.g.][in the context of chemical modelling]{Sahigara2013}, or local training data point density estimates \citep[e.g.][]{Aniceto2016}.
\item The strategy to compute AOA from DI; this is highly sensitive to the sampling design and the cross-validation strategy being used. The presented methodology accounts for that by deriving the threshold for the AOA from the DI in the training data where distances are calculated to data points that do not occur in the same cross validation fold. However, more systematic studies across various sampling designs and cross-validation strategies are needed to verify our assumption that the AOA represents the area for which, in average, the corresponding cross-validation error applies.
\item Related to this, the choice of the threshold to compute the AOA from DI; here we used the .95 quantile of the DI in the training data based on results of nearly 1000 model runs that all share a similar design. This recommendation comes with reservations, and a systematic analysis across various simulation model designs would provide a more robust measure. Predicting into unknown space will remain an uncertain adventure, and we believe that experiments across realistic simulation models may guide us into learning what this means.
\item The applicability across machine learning algorithms; here, the Random Forest algorithm was used. However, the problem of predicting beyond the data applies to other algorithms as well. Though not explicitly studied here, the approach should be applicable to other machine learning algorithms in the same way (if variable importance can be estimated) and is also not restricted to spatial data.
\end{itemize}
Hence, the results shown here should be considered as a baseline for ongoing discussions on this topic. 
The methodology to estimate the AOA has been implemented and published in the R package "CAST" \citep{Meyer2020}. The simulation studies are available as open source R scripts, and can be easily modified to obtain DI threshold values to derive AOA for other classes of simulation models and/or other spatial sampling designs. We strongly recommend researchers wanting to use the AOA to experiment with this.

%%%%%%%%%%%%%%%%%%%%%%%%%%%%%%%%%%%%%%%%%%%%%%%%%%%%%%%%%%%%%%%%%%%%
%%%%%%%%%%%%%%%%%%%%%%%%%%%%%%%%%%%%%%%%%%%%%%%%%%%%%%%%%%%%%%%%%%%%
\section{Conclusions}
We proposed a simple approach to map the area of applicability (AOA) of spatial prediction models, hence the area for which a model can be expected to make predictions with an expected error that is comparable to the cross-validation error of the model. Predictions outside the AOA should be handled with extreme care or be left out from further consideration because the environmental properties differ too strongly from those observed in the training data. Communicating the AOA is important to avoid mis-planning when predictive mapping is used as a tool for decision making (e.g. in the context of nature conservation), as well as to avoid propagation of massive errors when spatial predictions are used as input for subsequent modelling. We believe that the method proposed in this study will support critical assessment of overly optimistic data-driven prediction maps. We therefore suggest that the AOA should be provided alongside the prediction map and complementary to the communication of (cross-)validation performance measures. 

\section*{Data availability}
The methodology to estimate the AOA has been implemented and published in the R package "CAST" \citep{Meyer2020}. The simulation studies can be reproduced using the R-markdown scripts available from  \url{https://github.com/HannaMeyer/AOA_CaseStudy}.

\bibliography{Literature_AOA}

\end{document}